# Computer Vision based Tomography of Structures Using 3D Digital Image Correlation

Mehrdad S. Dizaji

Department of Mechanical Engineering, University of Massachusetts Lowell, 1 university Ave, Lowell, MA 01854, USA, Phone: (434) 987-9780; mehrdad_shafieidizaji@uml.edu

Devin K. Harris

Department of Engineering Systems and Environment, University of Virginia, 151 Engineer's Way, Charlottesville, VA, 22904, USA, Phone: (434) 924-6373; Email: dharris@virginia.edu

**Abstract and Introduction**

Internal properties of a sample can be observed by medical imaging tools, such as ultrasound devices, magnetic resonance imaging (MRI) and optical coherence tomography (OCT) which are based on relying on changes in material density or chemical composition [1-21]. As a preliminary investigation, the feasibility to detect interior defects inferred from the discrepancy in elasticity modulus distribution of a three-dimensional heterogeneous sample using only surface full-field measurements and finite element model updating as an inverse optimization algorithm without any assumption about local homogeneities and also the elasticity modulus distribution is investigated. Recently, the authors took advantages of the digital image correlation technique as a full field measurement in constitutive property identification of a full-scale steel component [22-27]. To the extension of previous works, in this brief technical note, the new idea intended at recovering unseen volumetric defect distributions within the interior of three-dimensional heterogeneous space of the structural component using 3D-Digital Image Correlation for structural identification [28-57]. As a proof of concept, the results of this paper illustrate the potential to identify invisible internal defect by the proposed computer vision technique establishes the potential for new opportunities to characterize internal heterogeneous materials for their mechanical property distribution and condition state.

**Research Significance**

The technique is proposed to discover material defects by way of property distribution analysis. In other words, analysis using the proposed technique can be used to detect both natural and induced defects such as voids, inclusions, impurities, contaminants, and other defects that may occur within a material and may therefore not be visible (i.e., visible with a naked eye because the defect is contained within the material) on the surface. These internal defects can be delineated and /or identified by their material properties distribution such as elastic modulus, density, or other material properties, as well as by physical properties such as the shape, size, and position. The finite element model updating process can include an iterative process in which external force is applied to boundary of a three-dimensional sample. Relying to the result of that initial iteration, more forces, less forces, more surfaces, less surfaces, or different areas of the previously tested surfaces can be applied to the sample to further determine if defects are present and where the defects are located. Using the identification of the existence and location of the defects, the defects can be eliminated from a sample or inspection process can be improved, for example, to add additional inspection steps.

Finite Element Model Updating problems generally tackles to determine for the non-homogeneous material property distribution, which requires knowledge of displacement fields and boundary conditions. Material properties, such as the young's modulus can be determined non-destructively. The mechanisms employed herein to create surface displacement and identify material properties of an internal area of a sample is referred to as a computer vision-based approach which takes advantages of heterogeneous property of strains on the surface of the structure to imply internal properties of the structure such as interior defects. This full field measurements data such as strains and displacements in different directions can be obtained using 3D Digital Image Correlation (3D-DIC) to generate displacement fields and collecting the displacement data on the surface resulting from those experiments and transforming that data using the proposed method to derive values from which to determine material properties and further characterize materials. Tools such as 3D-DIC have the potential to provide decision-makers with a comprehensive assessment tool to better describe the performance of the system while also being non-invasive and data-rich.

Traditionally, to simulate a structural component, for instance concrete components, a finite element model can be created, assuming the element is globally homogeneous. However, non-uniformly distribution of materials, interlayer fractures, and undesired porosity can be created during preparing the components which can cause them to have interior non-homogeneous properties. These internal non-homogeneous properties can be reflected on the boundary of the component by having particular form of strains/deformation patterns under external loadings. As an example, this concept of having different type of patterns for strain/displacement distribution is demonstrated in Figure 1, using two similar coupons tensile test, except one of them have some artificially damaged regions on the back side which cannot be seen from front. Two coupons tested in the same way. However, due to the damaged region on the back side of one of the coupons, the strain patterns are completely different. Leveraging 3D DIC technique as a full field measurement, and the particular strains/displacement patterns, internal heterogeneous status of the samples can be extracted locally, using interfacing initial finite element model with inverse optimization method and updating of structural system until discovering heterogeneous status distribution of the constitutive properties of the system.

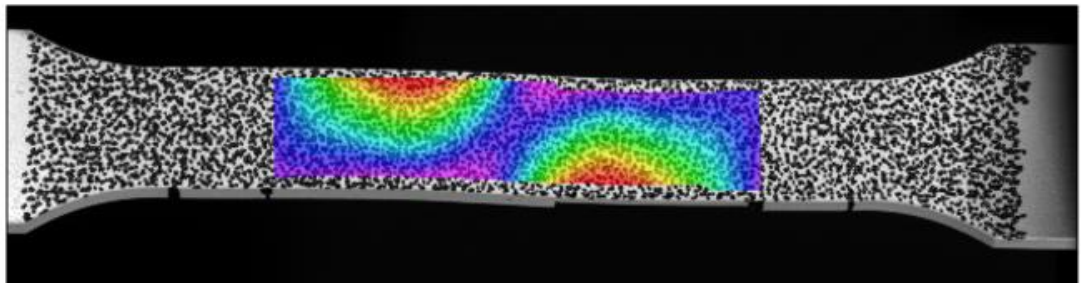
(a) Coupon with some artificially made damages on the back

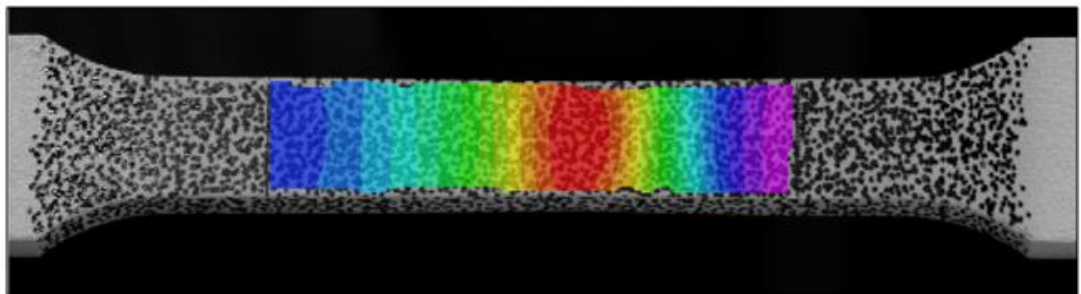
(b) Intact Coupon with no damages

*Figure 1. Different strain patterns for the coupons with and without damaged regions in the back side of the sample*

**Methodology**

The hypothesis of this work centers on the premise that internal defects can be delineated and/or inferred by their material constitutive properties distribution such as elastic modulus, shear modulus, density, or other material properties, and further described physical properties such as the shape, size, and position. The mechanisms employed herein to identify material properties of an internal area of a sample are informed by an image-based measurement approach which takes advantage of the heterogeneous characteristics of surface strains during loading to deduce internal properties (e.g., geometric features or defects) of the structure. Full-field surface deformation measurements derived from 3D Digital Image Correlation (3D DIC) have the potential to illustrate unseen anomalies within a solid body while also being non-invasive and data-rich.

Traditionally, to simulate a structural component, a finite element model can be created, assuming the element is globally homogeneous. However, non-uniform distribution of materials, (e.g., porosity, etc.), interlayer fractures, and defects are inherent to manufactured structural materials and can occur during the manufacture, resulting in a non-homogeneous internal property. For structural components with defects, the distribution of these internal non-homogeneous properties under loading (i.e., mechanical/thermal loadings) manifest in the form of perturbed strains/deformation patterns on the surface of the component. An example of the effects of non-uniform properties on strain/displacement patterns is demonstrated in Figure 2, using a tensile test of four similar coupons with different artificially manufactured defect features on the back side to mimic damaged regions, which are described as follows: (a) coupon specimen without any defect on the surface, subjected to tensile load, (b) coupon specimen with two artificially manufactured defects on the back side of the sample, which are invisible from the front side of the specimen subjected to tensile load, (c) coupon specimen with one artificially manufactured defect on the back side on the top region of the sample subjected to tensile load, and (d) coupon specimen with one artificially manufactured defect on the back side on the middle region of the sample subjected to tensile load.

During measurement, all of the coupons were tested in the same way; nonetheless, due to the damaged features on the back side of the coupons in configurations 2-4, the surface full-field strain and displacement patterns in different directions were clearly differentiable between the four represented specimens at the same load level. The non-homogeneous full-field strain/displacement patterns on the surface of the coupons can be inferred as a hologram of interior information such as existence of internal anomalies (e.g., defects). Therefore, in the proposed approach, an inverse problem is utilized to interpolate and tune those non-homogeneous surface patterns on the corresponding full-field strain/displacement surface pattern from numerical model by adjusting the variables (e.g., constitutive properties, boundary conditions or geometric properties) to infer internal properties (e.g., internal defects). Extracting more information from the surface of a sample can help to better imply and interpret the interior properties.

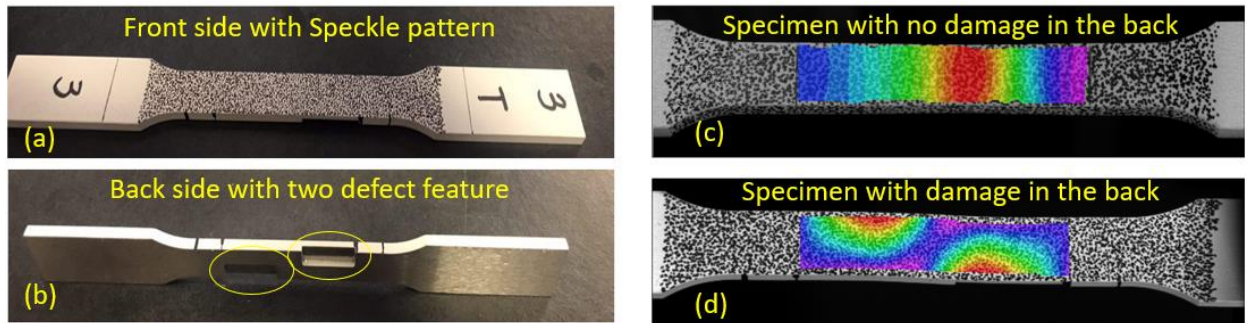

*Figure 2 Illustration of different specimens with DIC patterns (a) front side of coupon specimen with damage in the back, (b) back side of the coupon specimen with two rectangular damage, (c) longitudinal full-field strain pattern for the specimen without any damage in the back, and (d) longitudinal full-field strain pattern for the specimen with damages in the back side*

By leveraging 3D-DIC as a full-field measurement technique, and the observed strain/displacement patterns, the non-homogeneous internal status of the samples can be extracted locally and globally, by interfacing the initial finite element model with the measurements and updating the structural model until convergence. The proposed method is described as an image-based tomography of a structural components using an inverse approach.

In the Figure 3, the process of finite element model updating to establish projected damage evolution is depicted conceptually. During optimization process, the properties of each small partitions, as an ultimate goal, can be selected as design unknown parameters which can be modified iteratively, based on the optimization algorithm. For each time frame, the results from 3D DIC and FEA can be interpolated on each other and then the error, which is defined as residuals between 3D DIC and FEA results, will be calculated. Then, the design parameters will be adjusted based on the optimization algorithm until the results of DIC and FEA are tuned and correlated on each other. The geometric features (e.g., the features which are belong to internal and external defects) of the numerical simulation are recognized in the end of the finite element model updating process.

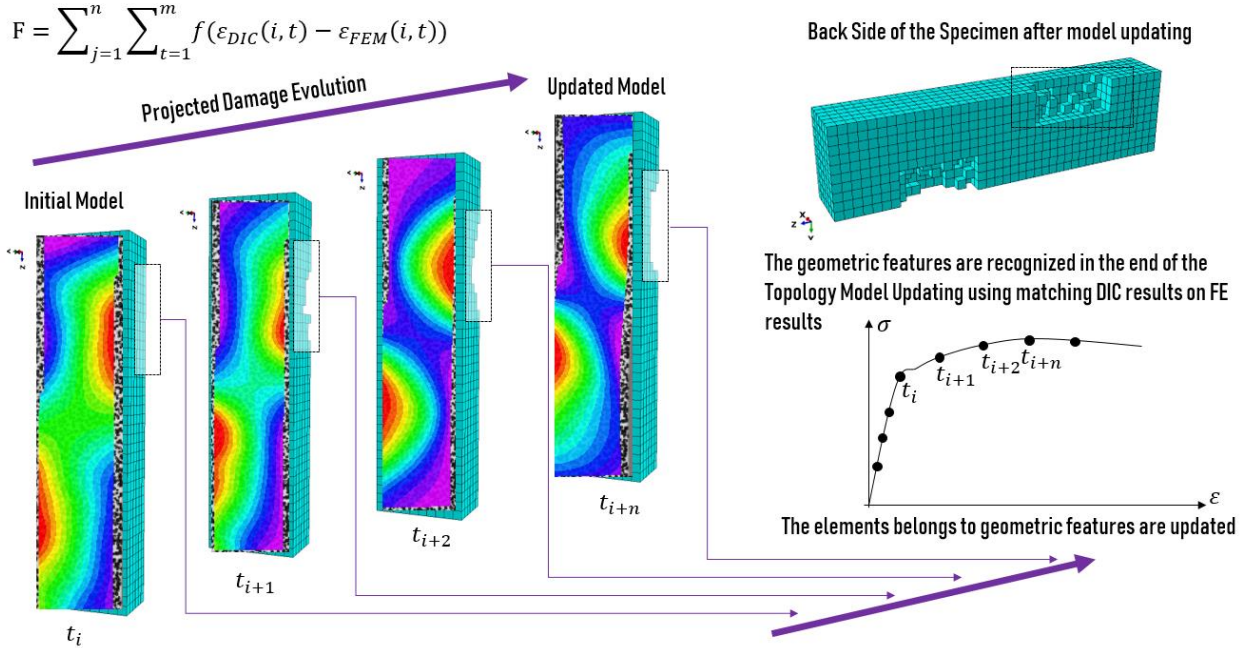

*Figure 3. Finite Element Model Updating process using DIC results*

This methodology does not require any prior knowledge about the problem domain. The techniques here are based on optimization method and finite element techniques as well as the elasticity modulus distribution represented as unknown design parameters on the mesh elements. The number of unknown design parameters (i.e., elasticity modulus values) are equal to the total of partitions. The basic principle of FEMU is to minimize the residuals between the experimentally measured and numerically computed response by adapting the unknown parameters of the finite element model. In this investigation a hybridized minimization algorithm is utilized to minimize the following cost function (Equation 1):

$$F(\boldsymbol{E}) = \sum_{i=1}^{m}\sum_{j=1}^{n_i}\left[\left(\frac{\varepsilon_{xx,ij}^{exp}-\varepsilon_{xx,ij}^{num}}{\varepsilon_{xx,ij}^{exp}}\right)^2 + \left(\frac{\varepsilon_{yy,ij}^{exp}-\varepsilon_{yy,ij}^{num}}{\varepsilon_{yy,ij}^{exp}}\right)^2 + \left(\frac{\varepsilon_{xy,ij}^{exp}-\varepsilon_{xy,ij}^{num}}{\varepsilon_{xy,ij}^{exp}}\right)^2\right] \quad (1)$$

With $\boldsymbol{E}$ the vector of unknown design parameters which are constitutive properties of the finite elements, $\boldsymbol{m}$ the number of load steps ($\boldsymbol{m} = 1$ in this work) and $\boldsymbol{n_i}$ the number of data points in the DIC measurement at load step $\boldsymbol{i}$. the subscripts *exp* and *num* indicate experimental response and numerical response respectively. The $\boldsymbol{\varepsilon_{xx}^{exp}}, \boldsymbol{\varepsilon_{yy}^{exp}}$ and $\boldsymbol{\varepsilon_{xy}^{exp}}$ represent the three components of the strain tensor, respectively that are extracted at a point $i$ of coordinates $x_i$ at time $t$. The values $\boldsymbol{\varepsilon_{xx}^{num}}, \boldsymbol{\varepsilon_{yy}^{num}}$ and $\boldsymbol{\varepsilon_{xy}^{num}}$ represent the corresponding values computed from the finite element model.

**Experiment setups and optimization process**

In this investigation, an experimental program was developed to evaluate the feasibility of leveraging 3D-DIC in a finite element model updating process to detect internal feature of a structural components or systems. The experimental program included to a laboratory scale investigation of a representative coupon sample subjected to displacement control loading. The structural configuration used in this investigation are illustrated schematically in Figure 4a. The tensile testing was coupon specimen carried out according to ASTM E8. The mechanical response of the specimen measured by 3D Digital Image Correlation (3D-DIC), a non-contact optical measurement technique capable of resolving surface deformations including strain and displacement.

As previously noted, finite element model updating process requires the development of an initial numerical model that can be updated based on experimentally derived behavior results. In this investigation, finite element model of the sample was developed in ABAQUS solver. It should be noted that ABAQUS allowed for the development of a direct interface with MATLAB, a multi paradigm numerical computing environment, which facilitated the iterative parameter optimization algorithm.

In the selected Hybrid Optimization Algorithm (HOA), first a genetic optimization step is employed to explore the space of parameters and locate the approximate region of the optimum solution. In the second step, a gradient-based method is utilized to continue the search within the approximate region to quickly converge on the precise location of the optimum solution. As a result, the favorable characteristics of both methods namely the efficient exploration of the space by the GA and the superior convergence of the gradient-based methods are leveraged to achieve an efficient optimization. As shown, a feasible initial guess for the parameters is used to start the process. The initial guess is used to generate an FEA model which upon analysis will be evaluated in the cost function. If the stopping criteria are not met, a new solution is generated through different operations in GA (e.g., selective reproduction, crossover and mutation). The new solution gives rise to a new FEA model and the process will be repeated as necessary. Once the stopping criteria are satisfied, the final solution of GA will be used to initiate the gradient-based scheme. This step will continue until convergence criteria are satisfied when the final optimal solution is identified. To conduct the investigation different artificial rectangular empty features are generated on the back side of the coupons which are going to realize as invisible interior defects (Figure 4b). To detect invisible damage, all FE elements can be considered within the updating process, as inputs into an objective function aimed at simultaneous local and global system parameter identification. However, in this experiment, as shown in the Figure 4b, instead of including all the elements in the process of the optimization, two rectangular empty feature, and nine sections of the coupon are inserted in the optimization process. The full-field measurements obtained from DIC and FEA, along with the absolute error, before and after model updating are shown in Figure 5a. As it can be noted from the Figure 5b, the initial values of the elasticity modulus are all different in the different region of the coupon before model updating but, as soon as finishing model updating process, the values of the elasticity modulus dropped dramatically in the notched locations, which are indication of damaged regions. Figure 4c illustrates a basic schematic of the HGA procedure adopted in this work.

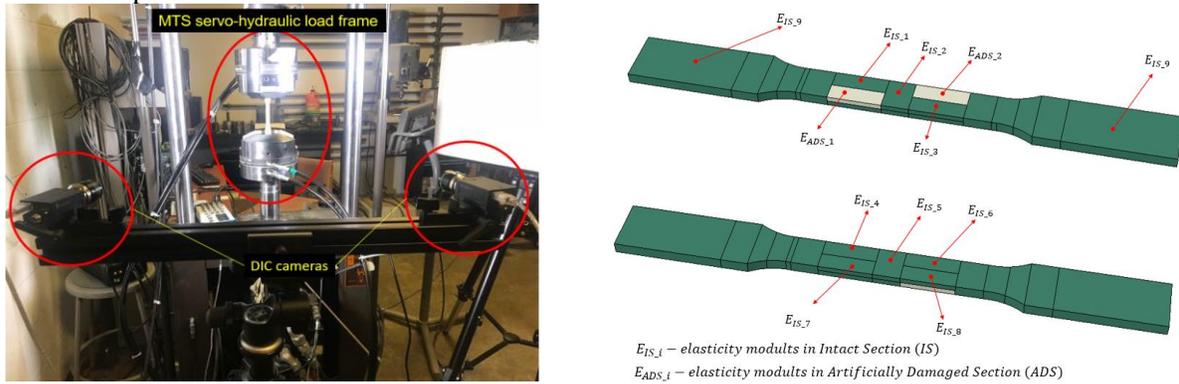

*Figure 4. (a) Experimental testing of validation specimen (b) Model of divided sections to import into optimization process*

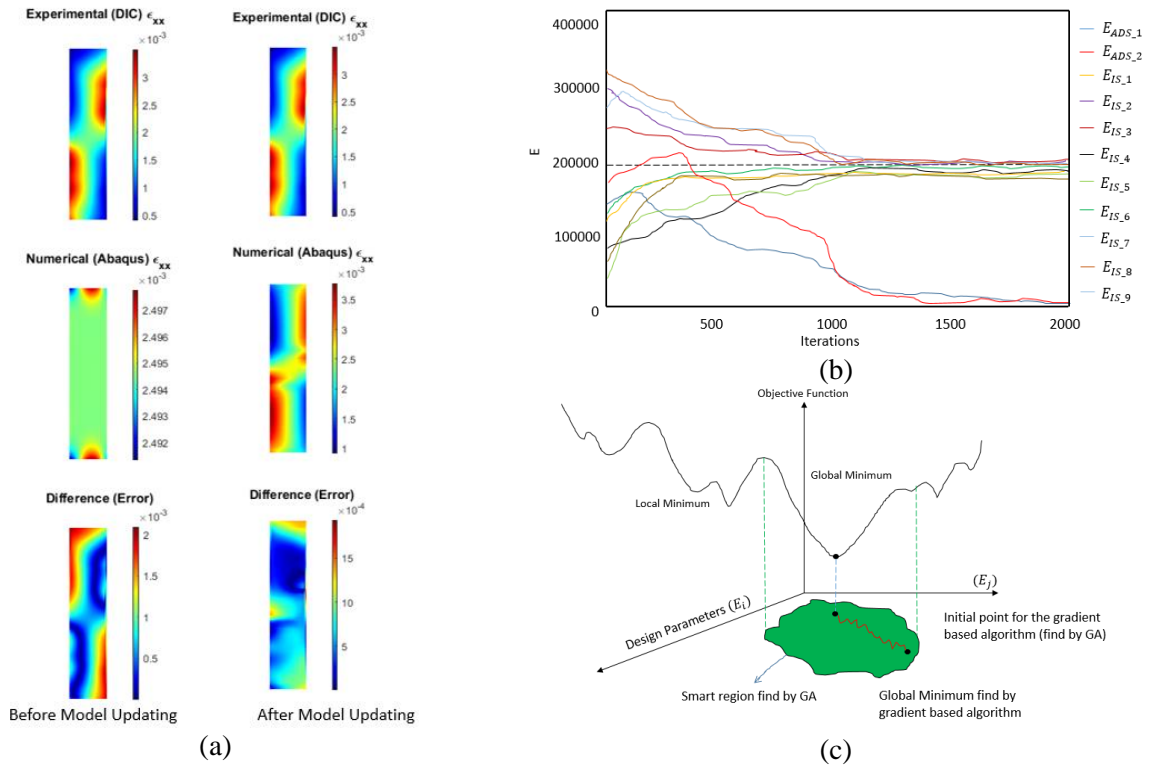

*Figure 5. (a) Full-Field measurements obtained from DIC and FE, along with the absolute error, before and after model updating (b) The convergence of elasticity modulus in different sections of the coupon (c) The proposed optimization algorithm is shown schematically*

**Summary**

The feasibility to recover the volumetric interior defect distribution, implied from discrepancy in elasticity modulus distribution, within interior space of the structural component, using solely surface measurements, without any assumption about local homogeneities and also the elasticity modulus distribution is studied for the first time. As a preliminary investigation, In the research aligned with structural identification work using 3D-DIC measurements, constitutive properties of small-scale element level validation are studied by conducting laboratory test on a steel coupon. The purpose of this investigation was to evaluate the capability of utilizing full-field measurements on structural identification to find constitutive properties of materials with the assumption that the coupon specimen has sections with different material properties. Digital Image Correlation technique is utilized to extract full-field measurement of the test sample.

This approach can not only determine material distribution of a sample that is homogenous or (intentionally) non-homogenous in its properties, but may also determine the location size, dimensions, and shape and to determine quantitative values for the material properties of defects that represent defects such as internal abnormalities, including those that may develop inside structural elements, or manufacturing defects that may also include voids or contaminants. Artificial synthetic defects, aiming to mimic an internal abnormality, are recovered within a coupon specimen using proposed optimization algorithm and finite element model updating as a reverse engineering method. To establish finite element model updating, ABAQUS solver interfaced with MATLAB as an optimization tool and unknown design parameters adjusted iteratively until finding optimal values. The results of this paper are encouraging and may open up new opportunities to characterize heterogeneous materials for their mechanical property distribution.